%% file: iclr2025_conference.tex
\documentclass{article} % For LaTeX2e
\usepackage{iclr2025_conference,times}

\input{math_commands.tex}

\usepackage[breaklinks]{hyperref}
\usepackage{url}
\usepackage{tabularx}
\usepackage{booktabs}
\usepackage{multirow}
\usepackage{graphicx}
\usepackage{wrapfig}
\usepackage{subcaption}

\title{The Impact of Model Zoo Size and Composition on Weight Space Learning}

\author{Damian Falk, Konstantin Schürholt, Damian Borth \\
AIML Lab,\\
School of Computer Science,\\
University of St.Gallen \\
\texttt{\{first.last\}$@$unisg.ch} \\
}

\iclrfinalcopy % Uncomment for camera-ready version, but NOT for submission.
\begin{document}

\maketitle
\begin{abstract}
Re-using trained neural network models is a common strategy to reduce training cost and transfer knowledge. 
Weight space learning - using the weights of trained models as data modality - is a promising new field to re-use populations of pre-trained models for future tasks. 
Approaches in this field have demonstrated high performance both on model analysis and weight generation tasks.
However, until now their learning setup requires homogeneous model zoos where all models share the same exact architecture, limiting their capability to generalize beyond the population of models they saw during training.
In this work, we remove this constraint and propose a modification to a common weight space learning method to accommodate training on heterogeneous populations of models. We further investigate the resulting impact of model diversity on generating unseen neural network model weights for zero-shot knowledge transfer.
Our extensive experimental evaluation shows that including models with varying underlying image datasets has a high impact on performance and generalization, for both in- and out-of-distribution settings. Code is available on \texttt{\href{https://github.com/HSG-AIML/MultiZoo-SANE}{github.com/HSG-AIML/MultiZoo-SANE}}.
\end{abstract}

\section{Introduction}
\label{sec:intro}

When training neural networks for computer vision applications, we follow a dominant paradigm of pre-training and fine-tuning ~\citep{panSurveyTransferLearning2010, yosinskiHowTransferableAre2014}, either by using pre-trained models trained from single datasets~\citep{mensinkFactorsInfluenceTransfer2021} or pre-trained foundation models, which can be used for fine-tuning to multiple downstream tasks~\citep{bommasani2021opportunities, qiu2024transferring}.

Given the vast amounts of pre-trained models, which have been deployed and released publicly on platforms such as Pytorch Hub or Huggingface, the research community has extended this paradigm by proposing the transfer or distillation of knowledge not only from one model but rather from a collection or population of pre-trained models.
These works can be categorized into training-based knowledge distillation methods~\citep{hintonDistillingKnowledgeNeural2015, lee2019learning, luo2020collaboration, jing2021amalgamating, yang2022factorizing}, where activation behavior or features are transferred, or training-free model merging~\citep{shuZooTuningAdaptiveTransfer2021, yang2022deepmodelreassembly, wortsmanModelSoupsAveraging2022, AinsworthHS23, xu2024training}, where model weights are aggregated given different heuristics. 

Recently, \textit{Weight Space Learning} has emerged as an additional approach to re-use populations of pre-trained models~\citep{schurholtSelfSupervisedRepresentationLearning2021, schurholtHyperRepresentationsGenerativeModels2022, navon2023equivariant, navon2023alignment, knyazev2023can, zhou2024permutation, schurholtScalableVersatileWeight2024, kofinasGraphNeuralNetworks2023, lim2024graph, meynentstructure}. This area of work could be categorized as training-based knowledge distillation done directly on model weights. 

Although training-based, weight space learning approaches do not need access to image datasets to create activation behavior as needed by training-based knowledge distillation methods. On the other side, being training-based, weight space learning methods might provide more adaptivity to unseen setups as training-free model merging techniques might be able to do. 

Weight space learning aims to learn a lower-dimensional representation of model weights given a population of models i.e., a model zoo. Such learned representations can be then exploited for multiple downstream tasks e.g., predicting the accuracy of neural networks directly from its weights or generating unseen neural network model weights.  
While previous work successfully demonstrated applications of weight space learning to the computer vision domain~\citep{schurholtSelfSupervisedRepresentationLearning2021,knyazev2023can,schurholtScalableVersatileWeight2024}, its scope was mostly limited to training representations on neural network models trained on the same image dataset e.g., CIFAR100. Such homogeneous \textit{single-zoo-training} setups neglect known benefits large and diverse pre-training datasets provide in machine learning~\citep{mensinkFactorsInfluenceTransfer2021, brownLanguageModelsAre2020, steinerHowTrainYour2021}.
To close this gap and motivated by the platonic representation hypothesis \citep{huh_platonic_2024}, which posits that representations learned by neural networks (NNs) converge, given sufficiently large model size and capacity, in this work, we investigate the effect of model zoo diversity on weight space learning beyond single-zoo-training using the SANE encoder-decoder backbone~\cite{schurholtScalableVersatileWeight2024}.
To that end, we identify sources of diversity in weight space learning as the model architecture, image dataset, and training hyperparameters of the underlying model zoo. We extend SANE training to a \textit{multi-zoo-setup}, where multiple model zoos trained on different image datasets are used for SANE backbone training. To make SANE suitable for non-homogeneous model zoo training, we adopt a novel per-token data normalization to enable and simplify data-processing for multiple model zoos at once.
We evaluate the proposed modification along two groups of model zoos: (i) a set of CNN model zoos representing smaller neural network architectures trained on $4$ different image datasets with in total 4000 model samples, and (ii) a set of ResNet models zoos representing larger neural network architectures trained on $3$ image datasets containing in total 3000 model samples. We test SANE's capability to zero-shot transfer knowledge in-distribution on model zoos that it already saw during training and out-of-distribution on models which it did not see during training. 
In both setups with the proposed modifications and suitable diversity, we outperform previous work and improve over single zoo training by on average $29.65$ and up to
$42.8$\% (CIFAR100 to EuroSAT) on ResNets, respectively.
\noindent
In summary, our contributions are as follows:
% \vspace{-12pt}
\begin{itemize}
    \item We extend SANE style weight space learning to accommodate pre-training on inhomogeneous model zoos.
    \item We identify axes for adding diversity as the model's architectures, datasets, and training hyper-parameters.
    \item We define an evaluation framework for analyzing the impact of diversity on weight space learning for both in- and out-of-distribution settings. 
    \item Using that framework, we systematically evaluate the performance impact of diversity in the pre-training data and model zoo size for different model sizes.
\end{itemize}

\section{Related Work}
\label{sec:related_work}
We structure this section according to the primary area of related work about weight space learning and the secondary area of related work about data diversity in pre-training.
\paragraph{Weight Space Learning}
Based on the observation that neural network weights become structured during training~\citep{martinTraditionalHeavyTailedSelf2019}, several approaches have been recently proposed to learn representations of model weights to make latent structure accessible:
by extracting high-information weight features to predict model properties ~\citep{eilertsenClassifyingClassifierDissecting2020,unterthiner2020predicting,martinPredictingTrendsQuality2021}, by training weight-decoders ~\citep{haHyperNetworks2017,zhangGraphHyperNetworksNeural2019,knyazev2021parameter,peeblesLearningLearnGenerative2022,knyazev2023can,wangNeuralNetworkParameter2024}, or as general encoder-decoder models for both tasks ~\citep{schurholtSelfSupervisedRepresentationLearning2021,schurholtHyperRepresentationsGenerativeModels2022,berardiLearningSpaceDeep2022,langoscoDetectingBackdoorsMetaModels2023,schurholtScalableVersatileWeight2024, meynentstructure}. 

% weight domain
In this context, several underlying learning backbones have been proposed ranging from simple MLPs
~\citep{eilertsenClassifyingClassifierDissecting2020,unterthiner2020predicting}, to CNNs ~\citep{berardiLearningSpaceDeep2022}, RNNs ~\citep{herrmannLearningUsefulRepresentations2024}, attention-based Transformers ~\citep{schurholtSelfSupervisedRepresentationLearning2021,schurholtHyperRepresentationsGenerativeModels2022,peeblesLearningLearnGenerative2022,andreisSetbasedNeuralNetwork2023,schurholtScalableVersatileWeight2024,soroDiffusionbasedNeuralNetwork2024a}, or Graph Neural Networks \citep{knyazev2021parameter,navon2023equivariant,kofinasGraphNeuralNetworks2023, zhouNeuralFunctionalTransformers2023,zhou2024permutation,lim2024graph, knyazevAcceleratingTrainingNeuron2024}.
% augmentations
In conjunction with backbone architectures, data augmentations have been proposed to improve generalization of weight space learning methods ~\citep{schurholtSelfSupervisedRepresentationLearning2021,shamsianImprovedGeneralizationWeight2024}.

To the best of our knowledge, this work is the first which aims at a \textit{multi-zoo-training} setup using an encoder-decoder architecture in weight space learning.

%\begin{itemize}
%    \item Learning Approaches
%    \item backbones
%    \item Augmentations
%    \item datasets
%\end{itemize}

%
%   knowlege transfer
%
\paragraph{Diversity in Knowledge Transfer}
Diversity of the underlying data plays a crucial role in transferring knowledge from source to target~\citep{mensinkFactorsInfluenceTransfer2021, shuZooTuningAdaptiveTransfer2021, you2022ranking, qiu2024transferring}. 
In particular, in setups where transfer is done from multiple sources or model zoos as in~\cite{shuZooTuningAdaptiveTransfer2021}, where more diverse training setups were able to outperform simple fine-tuning from a single pretrained model. 
In~\cite{you2022ranking} B-Tuning was proposed, which ranks multiple models given a model zoo according their suitability for finetuning. In experiments, the authors observed that knowledge transfer was consistently better when tuning with multiple models than a single one. However, in both works~\citep{shuZooTuningAdaptiveTransfer2021,you2022ranking}, a naive setup using all models from a model zoo does not necessarily yield best performance rendering the problem of selecting or combining the models non-trivial. Similar results have been reported in \cite{wortsmanModelSoupsAveraging2022}, where a linear combination or aggregation of model weights from a model yields improved results over single model performance of the zoo. In this work, a performance-based selection of models from the zoo is preferred over an aggregation of all models weights from the zoo. 
A different setup is outlined in~\cite{qiu2024transferring}, where the goal is to transfer knowledge from multiple foundation models to smaller downstream tasks models. For vision foundation models, the authors report a consistent out-performance of knowledge transfer from multiple foundation models over a single foundation model (independently of the underlying knowledge transfer approach).
Similar results have been reported in~\cite{rodriguez2024synergy}, where diverse variations of CLIP encoder models are combined and consistently outperform single CLIP models on a variety of underlying image datasets.
In both works, the effect is particularly visible in zero-shot scenarios.

\section{Methods}
\label{sec:methods}
In this section, we summarize the weight space learning we extend in this paper. Subsequently, we present an adaptation to make it suitable for inhomogeneous model weights.

\paragraph{Learning Backbone}
While there are various weight space learning methods, we base this work on encoder-decoder-based methods for their versatility. In particular, we extend SANE ~\cite{schurholtScalableVersatileWeight2024}. The core idea of SANE is to tokenize model weights and express entire models as sequences of token vectors. Using sequence models allows learning representations on chunks of the sequences, and still use the same SANE model on model sequences of different lengths, underlying architectures, and sizes. 

To that end, the model weights are reshaped into 2D matrices, then sliced into tokens $\mathbf{T}_n$ of size $d_t$. Zero padding or splitting is applied where needed to achieve same-size token vectors. For simplicity, we drop the sequence indices $n$ in the following. Each token is augmented by a 3-dimensional positional embedding, $\mathbf{P}=[n,l,k]$, to indicate sequence position $n$, layer index $l$, and within-layer position $k$. A binary mask $M$ distinguishes signal from padding.

The SANE model consists of an encoder $g_\theta$ that maps token sequences to sequences of token embeddings $\mathbf{z} = g_\theta(\mathbf{T},\mathbf{P})$, as well as a decoder $h_\psi$ that maps the token embedding sequence back to the original token space $\widehat{\mathbf{T}} = h_\psi(\mathbf{z},\mathbf{P})$. To structure the embedding space with a contrastive loss, a projection hat $p_\phi$ projects the latent embedding sequence to a lower dimensional space as $\mathbf{z}_p = p_\phi(\mathbf{z})$.

SANE is trained on chunks of token sequences with a combination of reconstruction and contrastive loss $\mathcal{L} = (1-\gamma) \mathcal{L}_{rec} + \gamma \mathcal{L}_c$:
\begin{align}
    \mathcal{L}_{rec} &=\| \mathbf{M} \odot \left( \mathbf{T}-\widehat{\mathbf{T}} \right)\|_2^2 \label{eq:seqhrep_recon_loss} \\
    \mathcal{L}_{c} &= NTXent(p_{\phi}(\mathbf{z_{i}}),p_{\phi}(\mathbf{z_{j}})). \label{eq:seqhrep_con_loss}
\end{align}
Here, the mask $\mathbf{M}$ indicates signal with $1$ and padding with $0$, to ensure that the loss is only computed on actual weights. The contrastive loss uses the augmented views $i,j$ and projection head $p_{\phi}$. \looseness-1

\paragraph{Masked Per-Token Loss Normalization}
Previous weight space learning work established that different weight distributions between different layers present a challenge for weight representation learning ~\citep{peeblesLearningLearnGenerative2022,schurholtHyperRepresentationsGenerativeModels2022,schurholtScalableVersatileWeight2024}. 
As remedies, they propose to either normalize the weights per layer across the entire dataset as a preprocessing step, or normalize the loss contribution accordingly. 
Both approaches present challenges for large, inhomogeneous weight datasets. They are not immediately applicable for varying architectures since they compute normalizations per layer and thus require matching architectures. Further, such normalizations may fail for different computer vision datasets with different weight distributions. Normalizing the loss per layer inherits these constraints and adds chunk-layer matching challenges if training is done on model chunks SANE-style. Therefore, since existing approaches do not work on zoos with inhomogeneous models, a new normalization mechanism is required to guide the backbone during training. 

Since normalizing the loss contribution is arguably more relevant for increased diversity, we therefore propose to normalize loss contributions \textit{per-token} at runtime. This has two benefits: (i) it simplifies the normalization and operates across different model architectures and weight distributions, (ii) the representation learning model still operates in weight space, which simplifies evaluating weight generation.

We standardize the target $T$ and prediction $\widehat{T}$ tokens as:
\begin{equation} T = \frac{T_{s,n} - T_\mu}{T_\sigma}, \quad \widehat{T} = \frac{\widehat{T_{s,n}} - T_\mu}{T_\sigma}, 
\label{eq:dnn_norm}
\end{equation}
where $T_\mu$ and $T_\sigma$ are the mean and std of the target token, respectively.
Depending on the architecture, SANE tokenization includes 0-padding to harmonize token size. Including the padding in the normalization would skew mean and std, usually towards zero. As an effect, this would overly increase the weight on tokens with more padding. To account for padding in the tokens, we normalize only on the signal as: 
\begin{align}
T_\mu &= \frac{1}{\sum_{i=1}^{N} M_i} \sum_{i=1}^{N} M_i \cdot T_{s,n}, \\
T_\sigma &= \sqrt{\frac{1}{\sum_{i=1}^{N} M_i} \sum_{i=1}^{N} M_i \cdot (T_{s,n} - T_\mu)^2} + \epsilon,
\label{eq:masked_mean_std}
\end{align}
where \( M_i \) is a binary mask that is 1 for valid elements and 0 for zero-padded elements, ensuring only valid data points contribute to the mean and standard deviation calculations.

%%%%%%%%%%%%%%%%%%%%%%%%%%%%%%%%%%%%%%%%
%
%%%%%%%%%%%%%%%%%%%%%%%%%%%%%%%%%%%%%%%%%

\section{Experiments}
\label{sec:experiments}
In this section, we test the proposed \textit{multi-zoo-training} setup and our hypothesis that increasing diversity in model weights can help transfer knowledge from the pre-training model population to out-of-distribution tasks. 
To that end, we first evaluate model weight averaging (also known as model souping) as baseline.
Subsequently, we turn to mutli-zoo SANE where we first evaluate the impact of per-token loss normalization. Subsequently, we use the differently trained SANE backbones to sample novel model weights and use the generated neural networks to evaluate their classification performance on the test split of different image datasets - either on an in-distribution (ID) or an out-of-distribution (OOD) image dataset.

\paragraph{Experiment Setup} 
For SANE training, we follow the experimental setup of ~\citet{schurholtScalableVersatileWeight2024}. 
As model zoo datasets, we use both small CNNs trained on MNIST~\citep{lecunGradientbasedLearningApplied1998}, SVHN~\citep{netzerReadingDigitsNatural2011}, USPS~\citep{hullDatabaseHandwrittenText1994}, and FMNIST~\citep{xiaoFashionMNISTNovelImage2017} as well as ResNet-18s trained on CIFAR10, CIFAR100~\citep{krizhevskyLearningMultipleLayers2009}, TinyImageNet~\citep{leTinyImageNetVisual2015}, SVHN, and EuroSAT~\citep{helber2019eurosat} from the model zoo dataset~\citep{schurholtModelZoosDataset2022}. Following previous work, we select models at epochs 21-25 for SANE training. We randomly split the models of the model zoo in train-validation-test splits of [70,15,15]. To pre-train SANE and sample models, we follow the training setup from ~\citet{schurholtScalableVersatileWeight2024}. The training parameters are summarized in Table \ref{tab:training_details} in App. \ref{app:experiment details}.

\subsection{Knowledge Aggregation via Model Souping}

To establish a baseline, we explore and evaluate a viable alternative before we continue with the proposed multi-zoo-training setup of the SANE backbone for knowledge transfer. 
Recently, merging models directly in weight space has gained attention. Different training-free methods have been proposed, averaging different training epochs of the same model~\citep{wortsmanRobustFinetuningZeroshot2022}, or averaging fine-tuned models that share a pre-trained model~\citep{wortsmanModelSoupsAveraging2022, rameModelRatatouilleRecycling2023}. Since populations of trained models do not generally share a single pre-trained model, an interesting approach is to re-align models before weight averaging. One such approach, git re-basin ~\citep{AinsworthHS23}, searches the permutation which changes the order of neurons per layer such that the weight distance between models is minimal. 

%%%%%%%%%%%%%%%%%%%%%%%%%%%%%%%%%%%%%%%%%%%%%%%%%%%%%%%%%%%%%%%%%%%%%%%%
\begin{wrapfigure}{r}{0.5\columnwidth}
\vspace{-14pt}
  \centering
    \includegraphics[width=0.5\columnwidth]{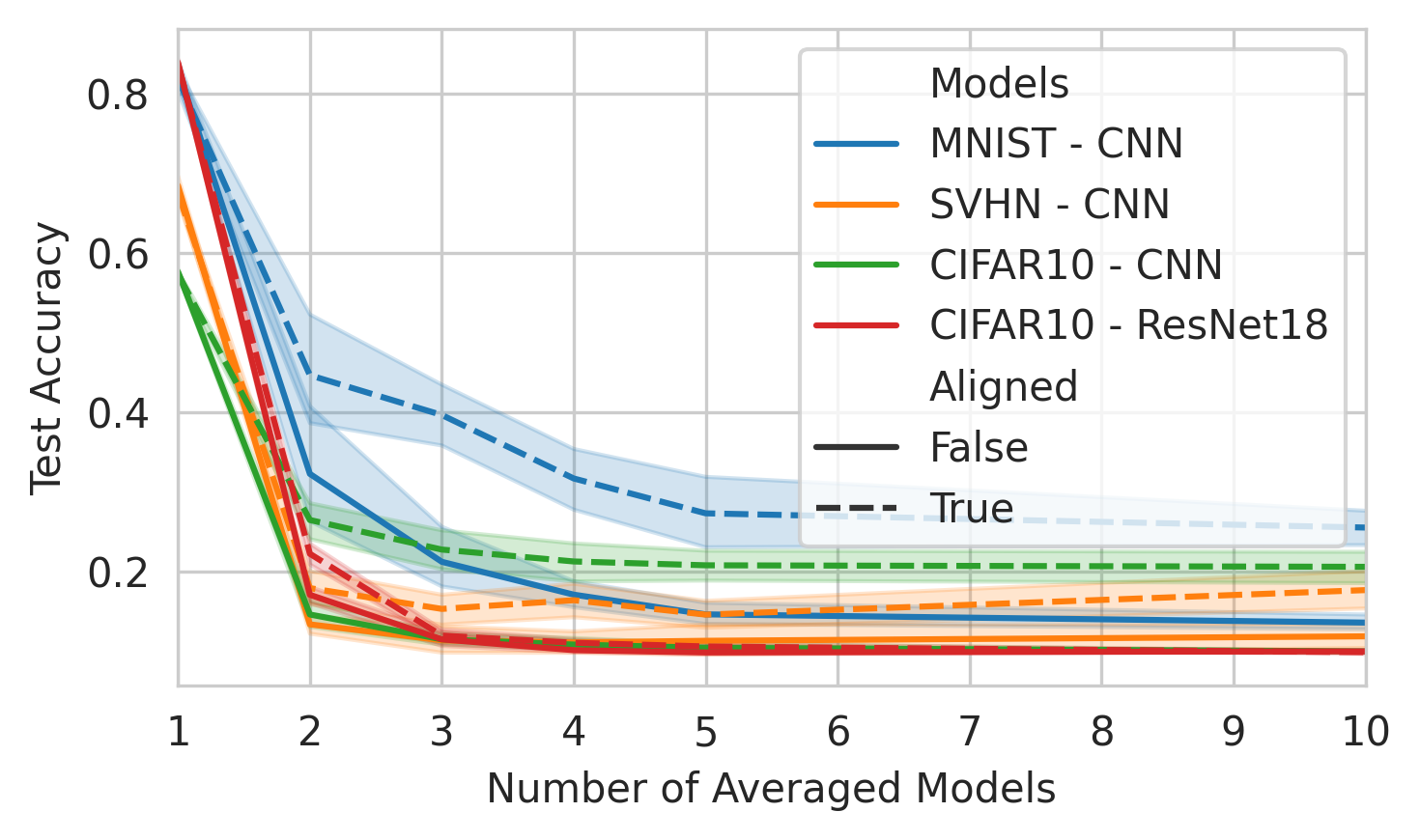}
\caption{Test accuracy of model soups over a number of averaged models. Increasing the number of models, aligned or not aligned, decreases performance. 
}
\label{fig:generative_soup_models}
\vspace{-8pt}
\end{wrapfigure}
%%%%%%%%%%%%%%%%%%%%%%%%%%%%%%%%%%%%%%%%%%%%%%%%%%%%%%%%%%%

To evaluate the suitability of weight-averaging models to aggregate knowledge, we therefore perform experiments on four model zoos, two with small CNN models, and two with larger ResNets. We randomly select models at epoch 25, average their weights, and evaluate their test performance on their original dataset. We evaluate averaging a varying number of models, with and without aligning, using git re-basin. 

Experimental evaluations on model soups with averaged weights show that weight averaging between different models is a challenging problem, as seen in Figure \ref{fig:generative_soup_models}. The performance of weight-averaged models decreases with the number of source models, compared to the single-model baseline. Aligning models generally improves performance over non-aligned source models, but only slightly.
 
Further, performance decreases with task and model complexity. Notably, even averaging aligned models decreases performance over the base population. This indicates that averaging weights of models that are not close to each other generally does not improve performance. While non-uniform weight averaging may improve the results, this indicates that new methods are needed for aggregation or knowledge transfer between populations of trained models. In the following, we evaluate SANE trained on multiple zoos.\looseness-1

\subsection{Masked Per-Token Loss Normalization}
\begin{figure}[h]
\vspace{-6pt}
  \centering
  \includegraphics[width=1.0\columnwidth]{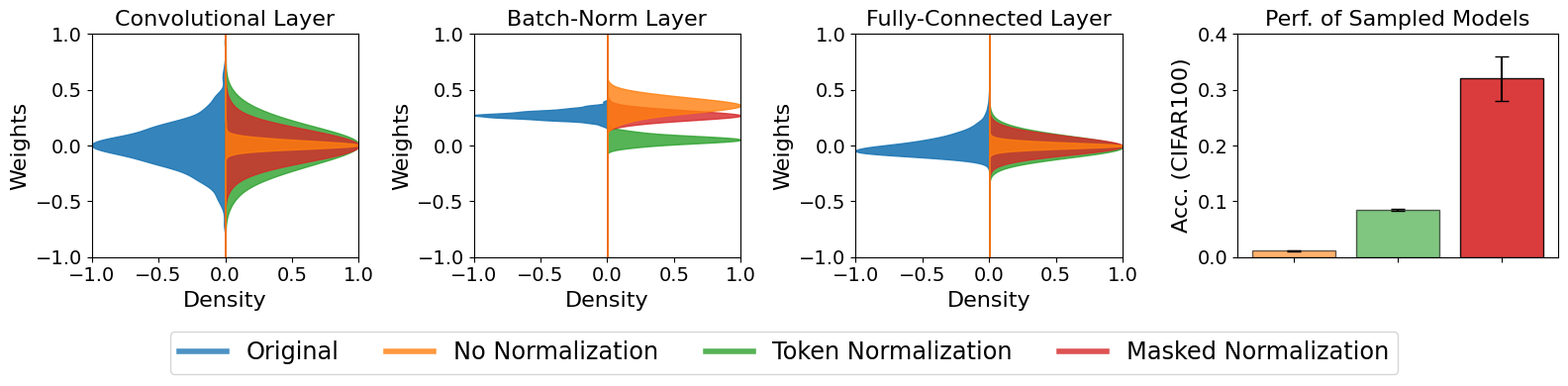}
  \caption{Comparison of weight distributions of a selection of ResNet layers between original weights (blue/left) vs reconstructed weights (right). We compare reconstruction without normalization (orange), with per-token normalization (green) and with masked per-token normalization (red). As in previous work, without normalization, weights of layers with narrow distributions are squashed towards the mean. Normalizing per-token fixes that issue. Ignoring the mask introduces a strong bias, particularly for batch-norm layers. Reconstructions with the masked per-token normalization match the original the closest. On the right we show the mean$\pm$std performance of 10 sampled ResNet-18 models on CIFAR100 with the different normalizations.}
  \label{fig:loss_norm}
\end{figure}

For the first experiment, we evaluate whether our extension to SANE enabling training on inhomogeneous zoos allows the SANE backbone to adequately capture the different weight distributions of different layers and models. This is crucial, since the encoder-decoder approach we are using as backbone operates in raw weight space, and skewed or squashed distributions - even of just individual layers - can have a catastrophic impact on the performance of generated models. To that end, we train SANE on CIFAR100 ResNet-18 models. Subsequently, we use models from the test split of the corresponding model zoos to be reconstructed by SANE (which corresponds to a simple forward pass through the encoder-decoder backbone). 

Following previous work, we use the match of weight distribution as a proxy for how well-reconstructed models mirror the original models ~\citep{schurholtHyperRepresentationsGenerativeModels2022}. Additionally we validate the results by sampling ResNet-18 models for CIFAR100 comparing the different normalization options. Results are shown in Figure \ref{fig:loss_norm}.

\paragraph{Loss normalization allows SANE training on inhomogeneous zoos} Our experiments demonstrate that training with per-token loss normalization allows training on inhomogeneous zoos without global weight normalization at dataset preprocessing time. Further, we did not encounter training instabilities, which might have been introduced for padding-heavy tokens. Lastly, masked loss normalization achieves a more accurate alignment between the reconstructed distribution and the original weight distribution across model parameters, see Figure \ref{fig:loss_norm}, particularly of batch-norm layer weights.
This alignment is especially pronounced in the larger ResNet-18 model zoos, where previous token-level normalization failed to capture the diverse weight behaviors accurately. By focusing on signal values only, the masked normalization more effectively maintains the original weight distributions, reducing reconstruction error and providing a stable signal even in high-parameter regimes.
These experiments confirm that SANE representations can be trained on inhomogeneous zoos with our masked per-token loss normalization. This allows us to evaluate the impact of different underlying computer vision datasets and other variations on knowledge transfer in the next section.

\subsection{Generating Models for Knowledge Transfer}

In the following, we evaluate whether sampling model weights with SANE to generate unseen neural networks can transfer knowledge from diverse populations. 
Specifically, we are interested if sampled models generalize better with increased diversity in SANE backbone pre-training given the proposed multi-zoo-training setup. Further, we are interested in the relation between the number of models of the corresponding model zoo used for SANE backbone training and the classification performance of generated models using the SANE backbone. %We perform experiments on both small CNNs as well ResNets to establish trends on different model sizes. 

\paragraph{Evaluation Criteria} 

Our emphasis for this work is to assess how well generating models using SANE can transfer knowledge from the pre-training model populations to the sampled models. To that end, we use the \textit{subsampling} weight generation method as introduced in~\cite{schurholtScalableVersatileWeight2024} using anchor samples  to account for the different architecture. Note that these models are generated by sampling in the latent of the learned representations and passed through the SANE decoder to generate an entirely new neural network model in a forward pass. In contrast to~\cite{schurholtScalableVersatileWeight2024}, no fine-tuning of the sampled models is done, which corresponds to the ``zero-shot'' setup.

%%% Place of table 1
\begin{table*}[h]
\centering
\caption{
Accuracy (mean $\pm$ std)  of sampled ResNet-18 models on the downstream image datasets. The single-zoo datasets each have 100 models with a total of 5M weight tokens each, while the multi-zoo dataset combines both having 200 models with a total of 10M weight tokens for training. }
\label{tab:eval_res_100_mz}
\begin{tabular}{ccccccc}
\toprule
\multicolumn{1}{c}{Single vs. Multi} & \multicolumn{2}{c}{In-Distribution} & \multicolumn{1}{c}{NOOD} & \multicolumn{2}{c}{FOOD} & AVG\\
\cmidrule(lr){1-1} \cmidrule(lr){2-3} \cmidrule(lr){4-4} \cmidrule(lr){5-6} \cmidrule(lr){7-7} 
Zoo                 & CIFAR10 & CIFAR100 &    TIN &   SVHN & EuroSAT &  \\
\cmidrule(lr){1-1} \cmidrule(lr){2-3} \cmidrule(lr){4-4} \cmidrule(lr){5-6} \cmidrule(lr){7-7}  
\multirow{1}{*}{CIFAR10} &    {30.2}$\pm${1.3} & {14.9}$\pm${0.8} &  {8.5}$\pm${0.4} & {18.9}$\pm${0.0} & {43.9}$\pm${1.4} &                   23.3$\pm$0.8 \\
\multirow{1}{*}{CIFAR100} &    {18.5}$\pm${0.6} &  {8.1}$\pm${0.4} &  {4.8}$\pm${0.4} & {21.3}$\pm${1.5} & {29.3}$\pm${2.9} &                   16.4$\pm$1.2 \\
\cmidrule(lr){1-1} \cmidrule(lr){2-3} \cmidrule(lr){4-4} \cmidrule(lr){5-6} \cmidrule(lr){7-7} 
\multirow{1}{*}{CIFAR10 + 100} &    \textbf{62.5}$\pm$\textbf{0.9} & \textbf{32.0}$\pm$\textbf{0.4} & \textbf{27.2}$\pm$\textbf{0.2} & \textbf{53.9}$\pm$\textbf{1.3} & \textbf{72.1}$\pm$\textbf{1.2} & \textbf{49.5}$\pm$\textbf{0.8} \\
\bottomrule
\end{tabular}
\end{table*}

We rigorously evaluate the generated models on in-distribution (ID) model zoos i.e., model zoos the SANE backbone was trained on, near-out-of-distribution (NOOD) and far out-of-distribution (FOOD) tasks i.e., model zoos which the SANE backbone did not see during training. We borrow the task relation from ~\cite{zhang_openood_2024} and detail the evaluation tasks in Table \ref{tab:model_zoo_selection}. 

%%%%%%%%%%%%%%%%%%%%%%%%%%%%%%%%%%%%%%%%%%%%%%%%%%%%%%%%%%%%%%

\paragraph{Increasing model zoo diversity during SANE training improves both ID and OOD performance}

\begin{wraptable}{r}{0.5\columnwidth}
\setlength{\tabcolsep}{4pt}
\centering
\vspace{-2pt}
\caption{Evaluation task classification for generated models. Models trained on some or all of the ID tasks form the pre-training model zoo for SANE. Models generated with SANE are systematically evaluated on in-distribution (ID) as well as corresponding near- (NOOD) and far out-of-distribution (FOOD) tasks. NOOD and FOOD terminology is borrowed from~\cite{zhang_openood_2024}.}
\small
\label{tab:model_zoo_selection}
\begin{tabularx}{0.5\columnwidth}{cccc}
\toprule
\textbf{Model Size} & \textbf{ID} & \textbf{NOOD} & \textbf{FOOD} \\
\cmidrule(lr){1-1} \cmidrule(lr){2-2} \cmidrule(lr){3-3} \cmidrule(lr){4-4}
\multirow{2}{*}{CNN} & \multirow{2}{*}{\shortstack{MNIST, \\ SVHN}} & \multirow{2}{*}{USPS} & \multirow{2}{*}{FMNIST} \\
                                      &                                              &                      &                          \\
\cmidrule(lr){1-1} \cmidrule(lr){2-2} \cmidrule(lr){3-3} \cmidrule(lr){4-4}
\multirow{3}{*}{ResNet-18} & \multirow{3}{*}{\shortstack{CIFAR10, \\ CIFAR100}} & \multirow{3}{*}{\shortstack{TinyImageNet \\ (TIN)}} & \multirow{3}{*}{\shortstack{SVHN, \\ EuroSAT}} \\
                                       &                                                     &                              &                          \\
                                       &                                                     &                              &                          \\
\bottomrule

\end{tabularx}
\vspace{-12pt}
\end{wraptable}

Transferring knowledge from multiple models trained on different datasets to a single target model is a challenging task. To test if SANE can be utilized for such scenarios, we first test the impact of the used model zoo for SANE backbone training on the performance of generated models and their classification performance on the corresponding image dataset. 

We therefore train SANE backbones in two setups: single-zoo and multi-zoo. In the single-zoos experiments, the model zoo dataset contains only models which have been trained on the same underlying image dataset, e.g. CIFAR10, while in a multi-zoo experiments we combine multiple model zoos together to form one larger and more diverse model zoo used for SANE backbone training.

We perform the experiments on both small CNNs ($\sim$2.5k params) to validate the method as well as larger ResNet-18 models ($\sim$12M params) to test if the method scales, following the evaluation scheme as outlined in Table \ref{tab:model_zoo_selection}. We focus on the ResNet-18 experiments in the paper and supplement the results on the smaller CNN zoos in App. \ref{sec:add_results}.

The results on the larger ResNet-18 model zoos as shown in Table \ref{tab:eval_res_100_mz} show a clear benefit of training on multiple zoos. SANE trained on both CIFAR10 and CIFAR100 models outperforms the single-zoo baselines across all metrics. Notably, the benefits are significant even over the respective ID training zoos, which suggests a positive knowledge transfer via SANE, even in ID experiments. The OOD evaluations show similar performance gains, which demonstrates that training SANE on models from multiple datasets allows to combine their knowledge for stronger OOD generalization.
What is more, the results outperform previous results from \cite{schurholtScalableVersatileWeight2024} for the more complex datasets by $\sim$ 15\% on TinyImageNet and $\sim$ 10\% on CIFAR100 while showing slightly lower performance on CIFAR10 ($\sim$ 5\% below SANE).

\begin{figure}[h]
  \centering
  \includegraphics[width=1.0\columnwidth]{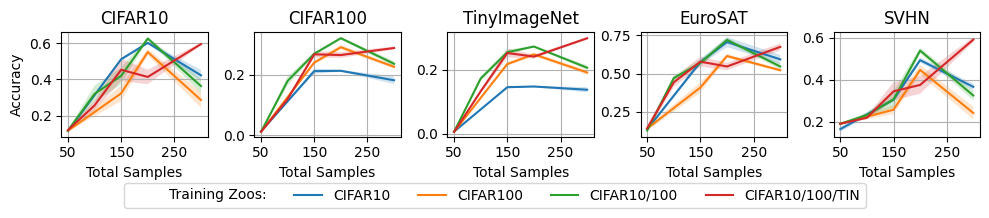}

  \caption{Comparison of 0-shot performance of sampled models on the downstream image datasets when varying model zoo composition and sample size. SANE is trained with [50/100/150/200/300] samples (2.5 - 15M weight tokens) for 60 epochs using data taken from one to three model zoos. }
  \label{fig:samp_div_comp}
  \vspace{-8pt}
\end{figure}

\paragraph{Training on more models improves transfer performance}

In our previous experiment, the size of the multi-zoo dataset is the combination of the two single-zoo datasets, and therefore has double the number of training samples. To evaluate how much impact the number of models has, we next evaluate with ResNet single-zoo and ResNet multi-zoo datasets with varying sample size and model zoo composition. When training the single-zoo baselines with the same sample size as the multi-zoo backbones, significant improvements of multi-zoo training can still be observed (for details see App. \ref{sec:add_results} Table \ref{tab:eval_res_200_mz} compared to Table \ref{tab:eval_res_100_mz}). However, the single-zoo performance improves significantly over the experiments with lower single-zoo sample size as well. This shows that the model zoo size used for backbone training has a large impact on the model generation performance. 

To further explore the relation between model zoo composition and sample size, we extend our experiments on ResNet-18s along both axes, training on one to three datasets (CIFAR10, CIFAR100, TIN) with varying sample size. The results are shown in Figure \ref{fig:samp_div_comp} and show that while model zoo size has a large impact on downstream performance irrespective of the number of model zoos used for training, important nuances can be observed. Increasing model count alone peaks earlier without further improving performance when adding more training samples, while increasing diversity without sufficient samples appears to undersample the more complex domain, leading to worse downstream performance. Interestingly, single-zoo backbones exhibit specific biases in OOD performance depending on class structure similarities. For example, a CIFAR10 trained backbone shows better performance on SVHN and EuroSAT, which share the same number of classes, whereas a CIFAR100 trained backbone excels on TinyImageNet but underperforms on EuroSAT and SVHN. In contrast, the multi-zoo backbones demonstrate a more balanced generalization, managing to perform reasonably well across both low- and high-class-count datasets. This suggests that diversity in training data supports broader adaptability and generalization across varying task complexities.

\paragraph{SANE initialized weights are amenable to further fine-tuning} Next, we compare our approach to a HyperNetwork \citep{haHyperNetworks2017} as an additional baseline. A key distinction is that HyperNetworks require image data to generate weights, while SANE learns purely from weight structure. This makes SANE’s generated weights effective initializations that remain amenable to fine-tuning. Therefore, SANE is orthogonal to and combinable with HyperNetworks or other data-driven weight generation methods. To validate this, we compare (i) SANE pre-trained on CIFAR10, CIFAR100 \& TIN with (ii) a HyperNetwork trained on CIFAR10, CIFAR100 \& TIN data, with task embeddings and architecture optimized as strong baseline, and (iii) random initialization as weak baseline.
We pre-train (i) and (ii), and fine-tune SANE sampled weights and HyperNetworks on the individual datasets.

\begin{figure}[h]
  \centering
  \includegraphics[width=1.0\columnwidth]{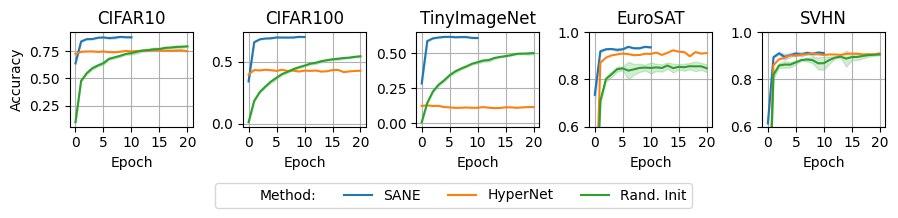}

  \caption{Comparison of SANE to HyperNetworks and random initialization during fine-tuning.}
  \label{fig:finetuning}
\end{figure}

Results (Fig. \ref{fig:finetuning}) demonstrate SANE's advantages: models initialized with SANE achieve both faster training and better final performance. While HyperNetworks perform reasonably well on OOD datasets, they show clear signs of overfitting during pre-training and gain no benefit from additional fine-tuning on the ID datasets despite hyperparameter optimization. 
This demonstrates that SANE's weight generation provides decent performance as is, and is also suitable for fine-tuning since it has not been trained on image data.

\section{Conclusion}
\label{sec:discussion}

In this paper, we evaluate the impact of variations in the weight training data on weight space learning. In particular, we evaluate an auto-encoder-based approach to weight-representation learning called SANE and evaluate the effect of combining model populations trained on different computer vision datasets. To facilitate this, we adapt SANE to handle heterogeneous model populations without prior weight normalization. Our experiments revealed that training SANE on diverse populations of models yields an intriguing effect: training on inhomogenous model zoos significantly enhances generalization when using enough training samples. Only increasing the sample size but not varying the composition of models used for training saturates earlier than when training on inhomogenous data. Single zoo baselines generalize well to OOD datasets with a similar number of classes but perform worse when sampling for datasets with more significant differences, indicating that multi-zoo training is a viable approach to improve generalization. As a baseline, we demonstrate that direct weight-averaging methods like model soups or git re-basin struggle to aggregate the knowledge of several models. Furthermore, other weight-generation methods such as HyperNetworks are prone to overfitting the training data and struggle to improve during finetuning. In contrast, SANE initialized models are also amenable to finetuning, since SANE does not use image data directly during training.\looseness-1

\bibliography{bib_auto, bib_manual}
\bibliographystyle{iclr2025_conference}

\appendix
\section{Appendix}

\subsection{Implementation Details}
\label{app:experiment details}

Experiments are performed in two phases: pre-training SANE on small CNN zoos and extending the analysis to larger ResNet-18 zoos. Model zoos are chosen based on their similarity in data and architecture to test SANE’s generalization ability across ID, NOOD, and FOOD domains. The hyperparameters used for training SANE are based on the SANE approach \citep{schurholtScalableVersatileWeight2024} and kept constant unless a modification is required to achieve stable training. They are summarized in Table \ref{tab:training_details}. Code to reproduce the experiments will be made available upon publication.

\begin{table}[h]
\centering
\caption{Architecture and hyperparameter choices for \textsc{SANE}. Unless otherwise specified, all experiments use the hyperparameters outlined below. The hyperparameters are based on the SANE approach \citep{schurholtScalableVersatileWeight2024} with modifications to the number of training epochs and learning rate to allow stable training with the proposed new loss normalization. }
\label{tab:training_details}
\begin{tabular}{lcc}
\toprule
\textbf{Hyper-Parameter} & \textbf{CNNs} & \textbf{ResNet-18} \\
\midrule
Tokensize ($T_\text{dim}$) & 289 & 288 \\
Sequence Length & $\sim$50 & $\sim$50k \\
Window Size ($W_s$) & 32 & 256 \\
Model Dim. ($D_\text{model}$) & 1024 & 2048 \\
Latent Dim. ($D_\text{lat}$) & 128 & 128 \\
Num. Transformer Layers & 4 & 8 \\
Num. Attention Heads & 8 & 8 \\
Num. Training Epochs & 50 & 60 \\
Learning Rate (LR) & $1e-4$ & $2e-5$ \\
Weight Decay (WD) & $3e-9$ & $3e-9$ \\
Batch Size & 32 & 32 \\
\bottomrule
\end{tabular}
\end{table}

\paragraph{Model Sampling} Model sampling is evaluated based on the performance of the sampled models on the downstream image dataset. All experiments use 5 prompt examples chosen from the model zoos at random to model the prior distribution out of which the decoder generates new weights. If there are no model zoos available, 5 models examples are trained for 25 epochs. For all experiments, the prompt example is chosen from the 25th epoch of training. Sampled models are evaluated without any updates of the trainable parameters (i.e. without any additional finetuning after sampling from SANE). Following the subsampling method, we sample 200 candidates and keep the 10 best models on validation data. As in SANE, batch-norm conditioning is performed before evaluation to update batch-norm statistics.

\subsection{Additional Results}
\label{sec:add_results}

\begin{table}[h]
\centering
\caption{Accuracy (mean $\pm$ std) of sampled CNN models on the downstream image datasets. The single-zoo datasets contain 200 models (10k weight tokens) each, the multi-zoo dataset is the combination of both and contains 400 models (20k weight tokens).}
\label{tab:eval_cnn_mz_top10_new}
\resizebox{0.8\columnwidth}{!}{%
\setlength{\tabcolsep}{2.5pt}
\begin{tabular}{ccccccc}
\toprule
\multicolumn{1}{c}{Single vs. Multi} & \multicolumn{2}{c}{In-Distribution} & \multicolumn{1}{c}{NOOD} & \multicolumn{1}{c}{FOOD} & \multicolumn{1}{c}{AVG} \\
\cmidrule(lr){1-1} \cmidrule(lr){2-3} \cmidrule(lr){4-4} \cmidrule(lr){5-5} \cmidrule(lr){6-6}
            Zoo       &           MNIST &            SVHN &            USPS &    FMNIST & \\
\cmidrule(lr){1-1} \cmidrule(lr){2-3} \cmidrule(lr){4-4} \cmidrule(lr){5-5} \cmidrule(lr){6-6}
\multirow{1}{*}{MNIST} & {84.7}$\pm${0.1} & {40.7}$\pm${2.4} & {52.0}$\pm${2.9} & {66.8}$\pm${0.1} &                   61.1$\pm$1.4 \\
\multirow{1}{*}{SVHN} & {83.4}$\pm${0.2} & {70.1}$\pm${0.1} & {68.0}$\pm${1.1} & {69.8}$\pm${0.1} &                   72.9$\pm$0.4 \\
\multirow{1}{*}{MNIST+SVHN} & \textbf{85.0}$\pm$\textbf{0.1} & \textbf{70.2}$\pm$\textbf{0.2} & \textbf{68.1}$\pm$\textbf{0.4} & \textbf{70.1}$\pm$\textbf{0.1} & \textbf{73.3}$\pm$\textbf{0.2} \\
\bottomrule
\end{tabular}
}
\end{table}

The experiments on CNN models show that combining different training zoos marginally outperforms the single-zoo baseline on the respective dataset, see Table \ref{tab:eval_cnn_mz_top10_new}. However, there are noticeable improvements on the other ID image datasets, compare, e.g., MNIST to SVHN. The results indicate that SANE backbone training on multiple zoos creates a superset of in-distribution datasets. 
The improvements are even more pronounced in OOD datasets. SANE backbone training on multiple model zoos (MNIST + SVHN), consistently outperforms single-zoo baselines across in-distribution (ID), near- (NOOD) and far out-of-distribution (FOOD) domains. 

\begin{table*}[h]
\centering
\caption{In contrast to results in Table \ref{tab:eval_res_100_mz}, this Table shows experiments, where we provide the same number of models (=tokens) for single-zoo-training and multi-zoo-training. SANE backbones are trained for each of the single zoo setups with 200 models (10M weight tokens), while the multi-zoo as combination is limited to 200 models (10M weight tokens). We report accuracy (mean $\pm$ std) of sampled ResNet-18 models on the downstream image datasets. }
\label{tab:eval_res_200_mz}
\begin{tabular}{ccccccc}
\toprule
\multicolumn{1}{c}{Single vs. Multi} & \multicolumn{2}{c}{In-Distribution} & \multicolumn{1}{c}{NOOD} & \multicolumn{2}{c}{FOOD} & \multicolumn{1}{c}{AVG} \\
\cmidrule(lr){1-1} \cmidrule(lr){2-3} \cmidrule(lr){4-4} \cmidrule(lr){5-6} \cmidrule(lr){7-7}
               Zoo & CIFAR10 & CIFAR100 &    TIN &   SVHN & EuroSAT \\
\cmidrule(lr){1-1} \cmidrule(lr){2-3} \cmidrule(lr){4-4} \cmidrule(lr){5-6} \cmidrule(lr){7-7}
 \multirow{1}{*}{CIFAR10}  & {60.1}$\pm${0.7} & {21.2}$\pm${0.3} & {14.7}$\pm${0.2} & {49.3}$\pm${1.2} & {70.7}$\pm${3.3} &                   43.2$\pm$1.1 \\
 \multirow{1}{*}{CIFAR100}  & {55.0}$\pm${1.8} & {29.0}$\pm${0.7} & {24.8}$\pm${0.5} & {44.7}$\pm${2.3} & {61.5}$\pm${1.3} &                   43.0$\pm$1.4 \\
\multirow{1}{*}{CIFAR10 + 100}  & \textbf{62.5}$\pm$\textbf{0.9} & \textbf{32.0}$\pm$\textbf{0.4} & \textbf{27.2}$\pm$\textbf{0.2} & \textbf{54.0}$\pm$\textbf{1.3} & \textbf{72.1}$\pm$\textbf{1.2} & \textbf{49.5}$\pm$\textbf{0.8} \\

\bottomrule
\end{tabular}
% }
\end{table*}

\end{document}

%% file: math_commands.tex
\usepackage{amsmath,amsfonts,bm}

\def\eqref#1{equation~\ref{#1}}
\def\1{\bm{1}}

\DeclareMathAlphabet{\mathsfit}{\encodingdefault}{\sfdefault}{m}{sl}
\SetMathAlphabet{\mathsfit}{bold}{\encodingdefault}{\sfdefault}{bx}{n}